\documentclass[journal]{IEEEtran}

\ifCLASSINFOpdf
\else
   \usepackage[dvips]{graphicx}
\fi
\usepackage{url}

\usepackage{graphicx}
\usepackage{times}
\usepackage{epsfig}
\usepackage{amsmath}
\usepackage{amssymb}

\usepackage{comment}
\usepackage{amsmath,amssymb} 
\usepackage{color}
\usepackage{caption}
\usepackage{bm}
\usepackage{multirow}
\usepackage{colortbl}

\usepackage[caption=false]{subfig}
\DeclareMathOperator{\softmax}{softmax}

\usepackage{float}
\newcommand{\minisection}[1]{\vspace{0.04in} \noindent {\bf #1}}
\usepackage{lineno}

\begin{document}

\title{On Implicit Attribute Localization for Generalized Zero-Shot Learning}

\author{Shiqi Yang, Kai Wang, Luis Herranz and Joost van de Weijer
\thanks{The authors \{syang, kwang, lherranz, joost@cvc.uab.es\} are with Computer Vision Center, Barcelona 08193, Spain. We acknowledge the support from Huawei Kirin Solution, the Spanish Government funding for projects PID2019-104174GB-I00 and RTI2018-102285-A-I00, and Kai acknowledges the Chinese Scholarship Council (CSC) No.201706170035. Luis acknowledges the Ramón y Cajal fellowship RYC2019-027020-I.}
}

\markboth{Journal of \LaTeX\ Class Files, Vol. 14, No. 8, August 2015}
{Shell \MakeLowercase{\textit{et al.}}: Bare Demo of IEEEtran.cls for IEEE Journals}
\maketitle

\begin{abstract}
Zero-shot learning (ZSL) aims to discriminate images from unseen classes by exploiting relations to seen classes via their attribute-based descriptions. Since attributes are often related to specific parts of objects, many recent works focus on discovering discriminative regions. However, these methods usually require additional complex part detection modules or attention mechanisms. In this paper,  1) we show that common ZSL backbones (without explicit attention nor part detection) can implicitly localize attributes, yet this property is not exploited. 2) Exploiting it, we then propose SELAR, a simple method that further encourages attribute localization, surprisingly achieving very competitive generalized ZSL (GZSL) performance when compared with more complex state-of-the-art methods. Our findings provide useful insight for designing future GZSL methods, and SELAR provides an easy to implement yet strong baseline.
\end{abstract}

\begin{IEEEkeywords}
Zero-shot learning, Attribute Localization
\end{IEEEkeywords}

\IEEEpeerreviewmaketitle
\vspace{-2mm}
\section{Introduction}
Visual classification with deep convolutional neural networks has achieved remarkable success~\cite{he2016deep,simonyan2014very}, even surpassing humans on some benchmarks~\cite{he2015delving}. This success, however, requires enough training images per class (tens to hundreds of images), which is often not the case in practice. The visual data to learn new classes may be scarce (i.e. few-shot learning) or inexistent (i.e. zero-shot learning -ZSL-). Humans, in contrast, are able to infer new classes from few or even no visual examples, just from a semantic description that connects them to known concepts
. Thus, ZSL is a desirable capability for computer vision systems, allowing them to recognize a larger set of classes via their semantic descriptions.

The most common representations in zero shot learning are global visual features extracted from a pretrained feature extractor, which are readily available off-the-shelf from previous works~\cite{Xian2018ZSLGBU,zhang2017learning}. These global visual features are then projected to a semantic space~\cite{Akata2016LabelEmbedding,Frome2013Devise} or to an intermediate space~\cite{zhang2015zero}, where the comparison with semantic representations takes place. In this paper, we focus on generalized zero-shot learning (GZSL), where the test set includes both seen and unseen classes. A major problem in GZSL is the model bias towards seen classes. 
Existing works can be roughly divided into two lines: generative and non-generative. Non-generative methods~\cite{Akata2016LabelEmbedding,Frome2013Devise,zhang2015zero,Xian2018ZSLGBU,zhang2017learning,elhoseiny2019creativity,zhu2018generative,Xie_2019_CVPR,zhu2019semantic,huynh2020fine}, \textcolor{black}{\cite{liu2020label, jiang2017learning,xie2020region}} focus on designing and learning a good visual-semantic alignment. Generative methods~\cite{xian2018feature,XianCVPR2019a,schonfeld2019generalized,wang2020bookworm,song2020generalized}, \textcolor{black}{\cite{shen2020invertible,gao2020zero}} adopt generative models to synthesize visual features of unseen classes. Generative methods have much higher performance than non-generative methods, since the classifier is trained with synthetic visual features of both seen and unseen classes (real and synthetic features, respectively), thus successfully avoiding the prediction bias towards seen classes. Note that generative methods \textcolor{black}{usually} need access to the descriptions of unseen classes during training (\textcolor{black}{however a counter-example is ~\cite{shen2020invertible}}), while non-generative GZSL methods generally assume they do not have access. Thus, a comparison between these two settings is not fair.

Most non-generative approaches focus on the role of the classifier and the semantic models, directly relying on global representations extracted by a pretrained network. The potential of local representations has been explored mainly recently in two directions: part detection~\cite{elhoseiny2019creativity,zhu2018generative}, and attention mechanisms~\cite{Xie_2019_CVPR,zhu2019semantic,huynh2020fine}. 
~\textcolor{black}{\cite{xie2020region} takes a step further, and combines part attention and leverages graph convolutional networks to reason about the parts relations.}
However, training part detectors normally requires additional and expensive annotation data (i.e. part ids, bounding boxes) to train the part detector. In contrast, attention mechanisms focus on discovering discriminative regions without requiring explicit region annotations. 
Both part detectors and attention mechanisms are significantly more complex and arguably more difficult to train than our proposed approach. Their representations are also essentially different to ours, since the attributes are not localized separately, but only a few regions are extracted. \textcolor{black}{Besides of those, \cite{jiang2017learning} tries to find a more discriminative latent attribute space, and \cite{liu2020label} learns another label space where labels of unseen classes can be regarded as linear combination of labels of seen classes, which is more suitable for the GZSL}. However, none of these methods notices the potential for attribute localization implicit in the feature extractor itself.

In this paper, we show a simple pipeline with fundamental modules is actually localizing attributes implicitly, a fact that is overlooked in literature. 
Then we show that the specific spatial aggregation method (implicitly, global average pooling - GAP - in most existing methods) becomes a critical choice in the design, since it can influence the degree of localization of the attributes. We then propose SELAR, a simple yet powerful variant of the previous baseline that uses global maximum pooling (GMP) as aggregation method. This simple modification further encourages attribute localization, boosting the performance and outperforming most GZSL methods with explicit attention. This suggests that implicit attribute localization provides better and more efficient representations than those learned with complex explicit attention modules.

\vspace{-2mm}
\section{Method}
\subsection{Task Definition}

In the GZSL, the training set contains seen classes and is defined as $\mathcal{S}\equiv \{(\mathbf{x}_i^s,y_i^s)\}_{i=1}^{N_s}$, where $i$ denotes the $i$-th image of the seen class and $y_i^s\in\mathcal{Y}^{\mathcal{S}}$ is its  class label. The test set contains unseen classes and is defined as $\mathcal{U}\equiv \{(\mathbf{x}_j^u,y_j^u)\}_{j=1}^{N_u}$. The sets of seen and unseen classes are disjoint, i.e.  $\mathcal{Y}^{\mathcal{S}}\cap\mathcal{Y}^{\mathcal{U}}=\emptyset$. The semantic information about a particular class $y$ is obtained by the class embedding function as $\psi\left(y_i\right)$. In the case of attribute-based representations with $L$ attributes, the class prototype $\psi_i=\psi\left(y_i\right)$ is simply a $L$-dimensional (binary or real valued) attribute vector
. In this way, the semantic information about all seen classes can be conveniently captured in a 
$\left|\mathcal{Y}^\mathcal{S}\right| \times L$-dimensional attribute matrix $A^\mathcal{S} \equiv \left[ \psi_1,\ldots,\psi_{\left|\mathcal{Y}^\mathcal{S}\right|} \right]^\intercal$.  Similarly, for unseen classes we obtain $A^\mathcal{U} \equiv \left[ \psi_1,\ldots,\psi_{\left|\mathcal{Y}^\mathcal{U}\right|} \right]^\intercal$. Finally, the evaluation in the GZSL setting considers a test set that includes both seen and unseen classes, i.e. $\mathcal{Y}^{\mathcal{SU}} =\mathcal{Y}^{\mathcal{S}}\cup\mathcal{Y}^{\mathcal{U}}$.

\begin{figure}[tbp]
	\centering
	{
		\includegraphics[scale=0.49]{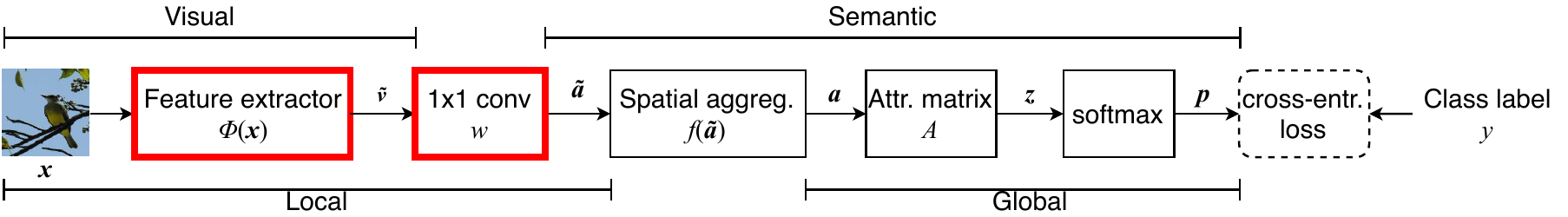}
	}\\
	\caption{A simple pipeline. Trainable modules are highlighted in red. The final fully connected layer is initialized with attribute matrix $A$ and fixed.\vspace{-2mm}}
	\vspace{-2mm}
	\label{fig:architectures}
\end{figure}

\begin{figure}[tbp]
	\centering
		\includegraphics[width=0.48\textwidth,height=0.15\textheight]{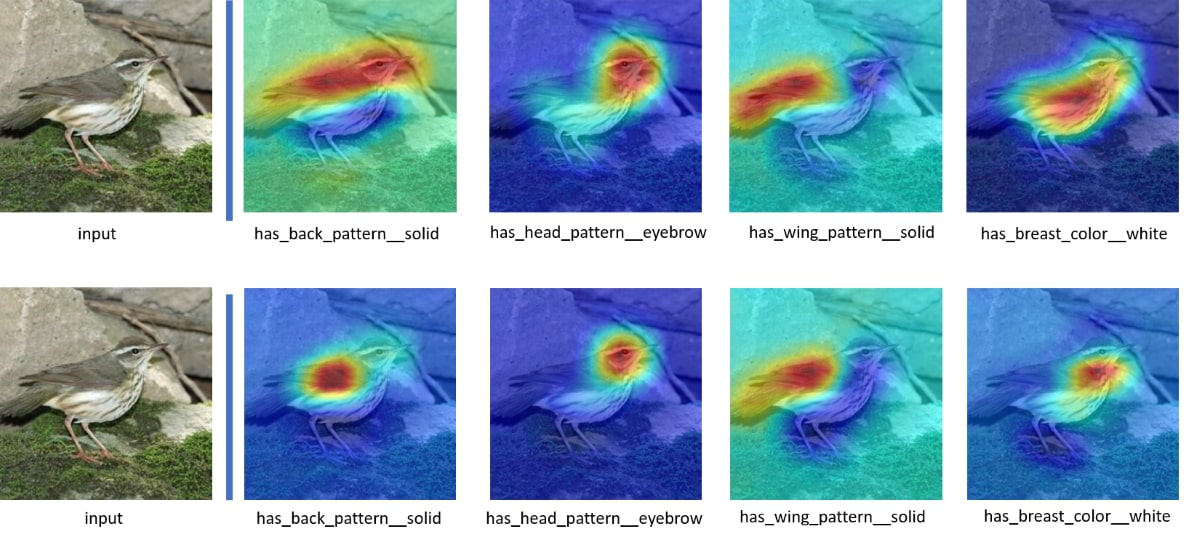}
	\caption{Attribute Activation Maps (AAMs) output by 1x1 conv from baseline (top) and SELAR (bottom) on CUB. 
	\vspace{-2mm}}
	\vspace{-2mm}
	\label{fig:aam}
\end{figure}

\begin{figure*}[tbp]
	\centering
	\subfloat[Baseline (seen)\label{fig:features_gap_seen}]{
		\includegraphics[width=0.23\textwidth,height=0.16\textheight]{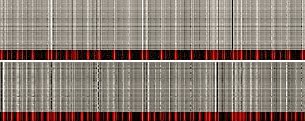}
	}
	\subfloat[Baseline (unseen)\label{fig:features_gap_unseen}]{
		\includegraphics[width=0.23\textwidth,height=0.16\textheight]{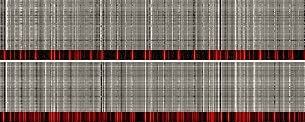}
	}
	\subfloat[SELAR (seen)\label{fig:features_gmp_seen}]{
		\includegraphics[width=0.23\textwidth,height=0.16\textheight]{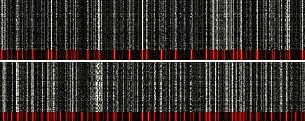}
	}
	\subfloat[SELAR (unseen)\label{fig:features_gmp_unseen}]{
		\includegraphics[width=0.2\textwidth,height=0.16\textheight]{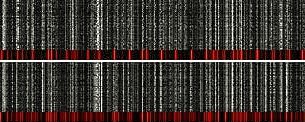}
	}\\
	\caption{Global semantic representations (rows) of 50 images per class (randomly selected) of 2 classes (super-rows). Each column corresponds to one of the 312 attributes. The description corresponding to the class (attribute vector) is shown in red (scaled to multiple rows). \textcolor{black}{\textbf{The ideal features (white one) should only have similar and sparse activation as the attribute vectors (red one)}.}
	\vspace{-4mm}}
	\vspace{-2mm}
	\label{fig:vis_em}
\end{figure*}

\vspace{-2mm}
\subsection{Implicit Attribute Localization in ZSL}

We formulate ZSL as a classification problem, using a deep convolutional neural network (CNN) that internally projects visual features to the semantic space and is trained end-to-end with cross-entropy loss on seen data. In particular, we are interested in certain intermediate representations: the \textit{local visual feature} $\mathbf{\tilde{v}}\in \mathbb{R}^{M\times M\times D}$, the \textit{global visual feature} $\mathbf{v}\in \mathbb{R}^D$ which is output by spatial aggregation, the \textit{global semantic feature} $\mathbf{a}\in \mathbb{R}^L$, and the logits or \textit{unnormalized class-scores} $\mathbf{z}\in \mathbb{R}^{\left|\mathcal{Y}^\mathcal{S}\right|}$. These intermediate representations lie in three distinctive spaces: the $D$-dimensional visual space, the $L$-dimensional semantic space (where $L$ is the number of attributes in our case) and the ${\left|\mathcal{Y}^\mathcal{S}\right|}$-dimensional class space.
For convenience, we can split the deep network into several modules: the \textit{feature extractor} $\mathbf{\tilde{v}}=\phi\left(\mathbf{x}\right)$, the \textit{spatial aggregation} operation $\mathbf{v}=f\left(\mathbf{\tilde{v}}\right)$, and the linear projection to the semantic space $\mathbf{a}=W\mathbf{v}$, parametrized by the embedding (or mapping) matrix $W\in \mathbb{R}^{L\times D}$, i.e. a fully connected layer. The embedding matrix is trainable from scratch, while the feature extractor is usually pretrained and can be optionally fine-tuned.
Finally, the overall loss to minimize is
\begin{equation}
\mathcal{L}=\mathbb{E}_{\left(\mathbf{x},y\right)\sim\mathcal{S}}\left[\mathcal{CE}\left(\softmax\left(A^\mathcal{S}Wf\left(\phi\left(\mathbf{x}\right)\right)\right),y\right)\right]
\end{equation}
where $\mathcal{CE}$ is the cross-entropy loss. The $A^\mathcal{S}Wf(\phi(\mathbf{x}))$ is known as bilinear compatibility score. It measures the similarity between the attribute vector and embedded visual features. 

Normally in ZSL, the feature extractor is a pretrained VGG~\cite{simonyan2014very} or ResNet~\cite{he2016deep} network, where spatial aggregation is global average pooling (GAP). Since both are linear operations, we can switch the order of GAP and the linear layer $W$ (which now becomes a $1\times 1$ convolution), as shown in Fig.~\ref{fig:architectures}. We emphasize that this simple pipeline (denoted as \textit{baseline}) already includes the key elements of ZSL methods operating on local features, that is, feature extractor (with the spatial aggregation) and embedding/mapping layer, since most methods at least contain these modules.

Since the baseline in Fig.~\ref{fig:architectures} (with GAP) is identical to a normal classification pipeline, we can get the Class Activation Map (CAM)~\cite{zhou2016learning}. The CAM is computed as:
\begin{equation}
C_c = \sum_{i=1}^{N_{att}} A_{c,i} \cdot \tilde{a}_i
\label{cam}
\end{equation}
where the $C_c$ is the CAM of class \emph{c}. It shows the most contributing region for the current class prediction. $N_{att}$ is the number of attributes, and $A_{c,i}$ means the $i$-th value in the attribute vector $A_c$. And the $\tilde{a}_i$ is the $i$-th channel of the feature map which is output by the feature extractor in Fig.~\ref{fig:architectures}.

CAM shows which part of the image contributes more to the current class prediction. The attribute vector $A_{c}$ gives the importance of all attributes for the class $c$. It actually weighs the different feature maps $\tilde{a}_i$ to achieve object localization. It is reasonable to conjecture that each feature map $\tilde{a}_i$ is related to a specific attribute (the $i-$th attribute in the attribute vector). When we visualize the four feature maps (referred to as Attribute Activation Map, AAM) output by the feature extractor from the baseline, as shown in the upper part of Fig.~\ref{fig:aam}, those four AAMs correspond to the four attribute which have the highest values in the attribute vector, \textit{i.e.}, having a higher contribution to the current class. We can observe that the AAMs are quite accurate performing weakly supervised attribute localization. Thus, the visual-semantic embedding is actually implicitly localizing the attribute regions.

\vspace{-2mm}
\subsection{SELAR: a simple yet effective method}
\label{spatial aggregation}

In Figs.~\ref{fig:features_gap_seen} and \ref{fig:features_gap_unseen}, we first visualize the embedded visual vectors (for both seen and unseen classes) from the baseline on the CUB dataset (312 dimension corresponding to 312 attributes) for 300 images (50 images per class). Each row is one embedded visual vector. We put together those vectors from the same class and join them with the attribute vector of this specific class (depicted in red). Because those features are already embedded into the semantic space, they are expected to be similar to the attribute vector. Ideally, the discriminative features are expected to have high values only for those attributes with high values in the ground-truth class attribute vector. But observing Figs.~\ref{fig:features_gap_seen} and \ref{fig:features_gap_unseen}, embedded visual feature vectors are also activated for some non-existing attributes (the ones which are not red in the attribute vector). 

We posit that this is due to the fact that GAP aggregates all spatial information of the feature maps (i.e. AAMs). Each spatial location in AAMs already corresponds to a large receptive field, while most attributes are typically  localized in a particular region, thus producing noisy features from GAP. 

In order to make the embedded visual vector more sparse and discriminative, we focus on obtaining more accurate AAMs. 
Therefore, we replace the GAP with global max pooling (GMP). The rationale behind this choice is that GMP has to focus on a single spatial location of the input AAMs (instead of averaging them as in GAP), thus enforcing the model to learn AAMs that are more localized .
The framework is shown in Fig.~\ref{fig:architectures}. We refer to this approach as \emph{SELAR} (\textbf{S}imple and \textbf{E}ffective \textbf{L}ocalized \textbf{A}ttribute \textbf{R}epresentation). Compared with the baseline, SELAR produces AAMs which correspond to more relevant regions, as shown in Fig.~\ref{fig:aam}. And the visualization of embedded visual vector shown in Figs.~\ref{fig:features_gmp_seen} and \ref{fig:features_gmp_unseen} clearly shows that SELAR generates more sparse and discriminative feature vectors, resembling more the actual attribute vectors describing the class.

\vspace{-2mm}

\begin{table*}[tbp]
	\centering
	
	\linespread{1.0}
	\addtolength{\tabcolsep}{-2pt}
    \scalebox{1.15}{\resizebox{0.80\textwidth}{!}{%
		\begin{tabular}{l  c c c c| c c c c| c c cc|c}
			&  \multicolumn{4}{c}{\textbf{SUN}} & \multicolumn{4}{c}{\textbf{CUB}} &  \multicolumn{4}{c}{\textbf{AWA2}}  &  \\
			\textbf{Method}  & \textbf{U} & \textbf{S} & \textbf{H}& \textbf{S/U}  &\textbf{U} & \textbf{S}  & \textbf{H}& \textbf{S/U} & \textbf{U} & \textbf{S} & \textbf{H}& \textbf{S/U}&$\bar{H}$\\
			\hline
			\hline
			\multicolumn{14}{c}{ \textbf{\textcolor{black}{Non-generative methods} (Unseen class descriptions are unknown during training)}}\\
			\hline
			PSR~\cite{annadani2018preserving} \textit{w/o}& $20.8$ & $37.2$ & $26.7$  &$1.79$& $24.6$ & $54.3$ & $33.9$  &$2.21$& ${20.7}$ & $73.8$ & ${32.3}$  &$3.57$ &$31.0 $\\
			DCN~\cite{liu2018generalized} \textit{w/o}& $\bm{25.5}$ & $37.0$ & $\bm{30.2}$  &$1.45$& $28.4$ & $60.7$ & $38.7$  &$2.14$& $25.5$ & $84.2$ & $39.1$   &$3.30$&$36.0 $\\
			MIIR~\cite{cacheux2019modeling} \textit{w/o}& $22.0$ & $34.1$ & $26.7$  &$1.55$& $30.4$ & $65.8$ & $41.2$  &$2.16$& $17.6$ & $87.0$ & $28.9$   &$4.94$&$32.3 $\\

			AREN~\cite{Xie_2019_CVPR}& $19.0$ & $38.8$ & $25.5$  &$2.04$& $38.9$  & ${78.7}$ & $52.1$ &$2.02$& $17.5$ & ${93.2}$ & $29.5$  &$5.33$&$35.7 $ \\
			JLA~\cite{lijoint}& $23.2$ & $36.6$ & $28.4$  &$1.58$& $36.6$  & $59.8$ & $45.4$  &$1.63$& $24.5$ & $91.6$ & $38.3$  &$3.74$ &$37.4 $\\
			AttentionZSL~\cite{Liu_2019_ICCV}&  $18.5$ & $\bm{40.0}$ & $25.3$  &$2.16$& $36.2^*$  & $\bm{80.9}^*$ & $50.0^*$  &$2.23$& $27.0^*$ & $\bm{93.4}^*$ & $41.9^*$   &$3.46$&$39.1 $\\
		 
			SGMA~\cite{zhu2019semantic}&  $-$ & $-$ & $-$ &$-$ & $36.7^*$  & $71.3^*$ & $48.5^*$ &$1.94$ & $\bm{37.6^*}$ & $87.1^*$ & $\bm{52.5^*}$  &$\bm{2.32}$ &$-$\\ 

			\hline
			\textbf{Baseline \textit{w/o}}& ${23.8}$ & $32.0$ & ${27.3}$  &$\bm{1.34}$& ${32.1}$  & $63.0$ & ${42.5}$  &$1.96$& $12.0 $ & ${87.2} $ & $21.0 $  &$7.27$&$30.3 $\\
			
			
			\textbf{Baseline}& ${23.4}$ & ${37.2}$ & ${28.7}$  &$1.59$& $\textcolor{black}{39.0}(37.1^*)$  & $\textcolor{black}{74.2}(73.2^*)$ & $\textcolor{black}{51.1}({49.2^*})$  &$1.97$& $13.7(\textcolor{black}{14.6^*}) $ & $90.4(\textcolor{black}{77.0^*}) $ & $23.8(\textcolor{black}{24.5^*}) $  &$6.60$ &$ 33.9$\\
			
			\textbf{SELAR \textit{w/o}}&$22.8 $ & $31.6$ & $26.5$  &$1.39$& ${43.5}$  & ${71.2}$ & ${54.0}$  &${1.64}$& ${31.6} $ & $80.3 $ & ${45.3} $   &${2.54}$&${41.9} $\\
			
			\textbf{SELAR}&${23.8}$ & ${37.2}$ & ${29.0}$  &${1.56}$& $\textcolor{black}{43.0}(\bm{51.4}^*)$  & $\textcolor{black}{76.3}(75.2^*)$ & $\textcolor{black}{55.0}(\bm{61.0}^*)$ &$\bm{1.46}$ & ${32.9}(\textcolor{black}{29.5^*}) $ & $78.7(\textcolor{black}{80.2^*}) $ & ${46.4}(\textcolor{black}{43.2^*}) $  &${2.39}$&$\textbf{45.5}$\\
			
			\hline
			\hline
			\multicolumn{14}{c}{\textbf{\textcolor{black}{Generative methods} (Unseen class descriptions are known during training)}}\\
			\hline 
			f-CLSWGAN~\cite{xian2018feature}&  $42.6$ & $36.6$ & ${39.4}$ &$-$ & $43.7$  & $57.7$ & $49.7$ &$ -$ & $52.1$ & $68.9$ & ${59.4}$  &$- $ &$49.5$\\
			f-VAEGAN-D2~\cite{XianCVPR2019a}&  $45.1$ & ${38.0}$ & $41.3$ &$-$ & $48.4$  & $60.1$ & $53.6$ &$ -$ & ${57.6}$ & $70.6$ & ${63.5}$  &$- $ &$52.8$\\
			CADA-VAE~\cite{schonfeld2019generalized}&  $\bm{47.2}$ & $35.7$ & ${40.6}$ &$-$ & $51.6$  & $53.5$ & $52.4$ &$ -$ & $55.8$ & ${75.0}$ & ${63.9}$  &$ -$ &$52.3$\\
			attr-BImag~\cite{wang2020bookworm}& $22.4$ & $40.1$ & $29.2$ & $-$ & $41.3$ & $\bm{77.7}$ & $53.9$ & $-$ & $\bm{60.0}$ & $73.8$ & $\bm{66.2}$ & $-$ &$49.8$\\
			class-attr-BImag~\cite{wang2020bookworm}& $21.7 $& ${40.7}$ & $28.2$ & $-$ & $44.1$ & $73.6$ & $55.9$ & $-$ & $51.4$ & $\bm{76.9}$ & $61.6$ & $-$ & $48.6$\\
			\textcolor{black}{$\ddagger$IZF-NBC}~\cite{shen2020invertible}&  $44.5$ & $\bm{50.6}$ & $\bm{47.4}$  &$ $& $44.2$  & $56.3$ & $49.5$  &$ $& $58.1$ & $76.0$ & $65.9$   &$ $&$ \bm{54.3} $\\
			\hline
			\textcolor{black}{\textbf{$\dagger$SELAR}}&  $40.5$ & $32.9$ & $36.3$ &$-$ & $\bm{62.4}$  & $64.9$ & $\bm{63.6}$ &$ -$ & $52.0$ & $71.9$ & ${60.3}$  &$ -$ &$53.4$\\
			\hline
		\end{tabular} }
	}
	\caption{Generalized Zero-Shot Learning on Proposed Split (PS). \textbf{U} = Top-1 accuracy on $\mathcal{Y}_{\mathcal{U}}$, \textbf{S} = Top-1 accuracy on $\mathcal{Y}_{\mathcal{S}}$, H = harmonic mean, \textbf{S/U} can show the bias towards seen class, $\bar{H}$ denotes the average over the H on three datasets. * indicates results using VGG19 as feature extractor while others use ResNet-101. \textit{w/o} means not finetuning feature extractor. We highlight the best results for both \textcolor{black}{non-generative and generative methods}. $\dagger$ indicates using calibration, which assumes access to a small validation set of seen and unseen classes. $\ddagger$ means not requiring accessing to unseen class descriptions. \vspace{-2mm}}
	\label{tab2}
	\vspace{-2mm}
\end{table*}

\section{Experiments}\label{sec:experiments}

\minisection{Settings.} We evaluate our method on three datasets: CUB~\cite{WelinderEtal2010}, SUN~\cite{patterson2014sun} and AWA2~\cite{Xian2018ZSLGBU} under the challenging GZSL setting. We denote the accuracy on unseen classes and seen classes as $Acc_U$ and $Acc_S$, respectively, and the evaluation metric for GZSL is the harmonic mean, calculated as $H=2*Acc_U*Acc_S/(Acc_U+Acc_S)$. Our model is built on top of ImageNet pretrained model. We initialize the last fully connected layer with the L2-normalized attribute matrix, then remaining fixed. 
For fair comparison, we distinguish between methods that assume the descriptions of unseen classes are unknown during training (typically non-generative methods), and methods that assume they are known (generative methods).

\minisection{GZSL performance.} Table~\ref{tab2} shows the results. 
Regarding methods that do not assume knowledge about unseen classes, SELAR achieves state-of-the-art performance on CUB and competitive results on SUN and AWA2. Notably, it surpasses other more complex methods on the fine-grained CUB dataset significantly. We conjecture that this is likely due to the fact that attributes of CUB are very localized, and easily linked to visual patterns, unlike in AWA2 and SUN, which have abstract attributes such as \textit{smart} and \textit{domestic}. Among the methods in Table~\ref{tab2}, AREN~\cite{Xie_2019_CVPR}, JLA~\cite{lijoint} and AttentionZSL~\cite{Liu_2019_ICCV} utilize extra modules to localize parts or obtain attention maps. AREN also has one branch with explicit self-attention and GMP, and SGMA~\cite{zhu2019semantic} has additional part detection modules. However, these methods obtain inferior results on most datasets compared to our method, despite of their increased complexity. While SGMA achieves state-of-the-art performance on AWA2, however, it requires four forward passes through the feature extractor and larger input resolution. Specifically for SUN, using GAP or GMP does not make much difference. We posit that this is due to the fact that attributes in the SUN dataset are not always clearly localizable  (like the attribute \textit{natural light}); whether to consider all these regions by GAP or only a single location by GMP does not influence a lot. We also report the average $H$ over all these datasets (we do not include here methods with results only on two datasets) in Table~\ref{tab2}. For this metric SELAR has the highest value among all these methods. The ratio $S/U$ (only for non-generative and without calibration) suggests that our method has lower seen-unseen bias than others. In general, we observe that two simple modifications of the backbone network (i.e. mapping to localized semantic space with a $1\times 1$ convolution and encouraging more discriminative localization with GMP ) result in significant gains and lower bias, and surprisingly good results, especially considering that the other methods use much more complex architectures.

We also report results for SELAR with calibration stacking~\cite{chao2016empirical}. Calibration stacking helps to avoid bias towards seen classes by decreasing the prediction score for all seen classes during evaluation. However, finding the precise factor requires observing a small set of validation images (both seen and useen). In addition, for reference, we also include results of several generative models that assume knowledge (\textcolor{black}{except IZF-NBC~\cite{shen2020invertible}}) of the descriptions of unseen classes (i.e. images are unseen, but class descriptions are seen), which synthesize features for unseen data, and thus training a less biased classifier with seen and unseen classes.

\begin{table}[!tb]
	\centering
	
 	\resizebox{0.5\textwidth}{!}{%
	\begin{tabular}{cc|ccc}
		Type   & Space & U    & S    & H           \\
		\hline
		GAP       & visual, attribute, class (all equivalent)         & $37.1$ & $73.2$ & $49.2$        \\
		GMP        & visual         & $39.1$ & $ \bm{80.7}$ & $52.7$        \\
		GMP & attribute (ours)       & $ \bm{51.4}$ & $75.2$ & $ \bm{61.0} $       \\
		GMP &  class         & $26.3 $ & $74.7 $ & $38.9  $       \\
		\hline
	\end{tabular}
	}
	\caption{Ablation study of pooling operations and pooling spaces. The results are reported on the CUB dataset.
	\vspace{-4mm}}
	\label{table:localization}
\end{table}

\begin{figure}[tb]\centering
	\includegraphics[width=0.5\textwidth,height=0.18\textheight]{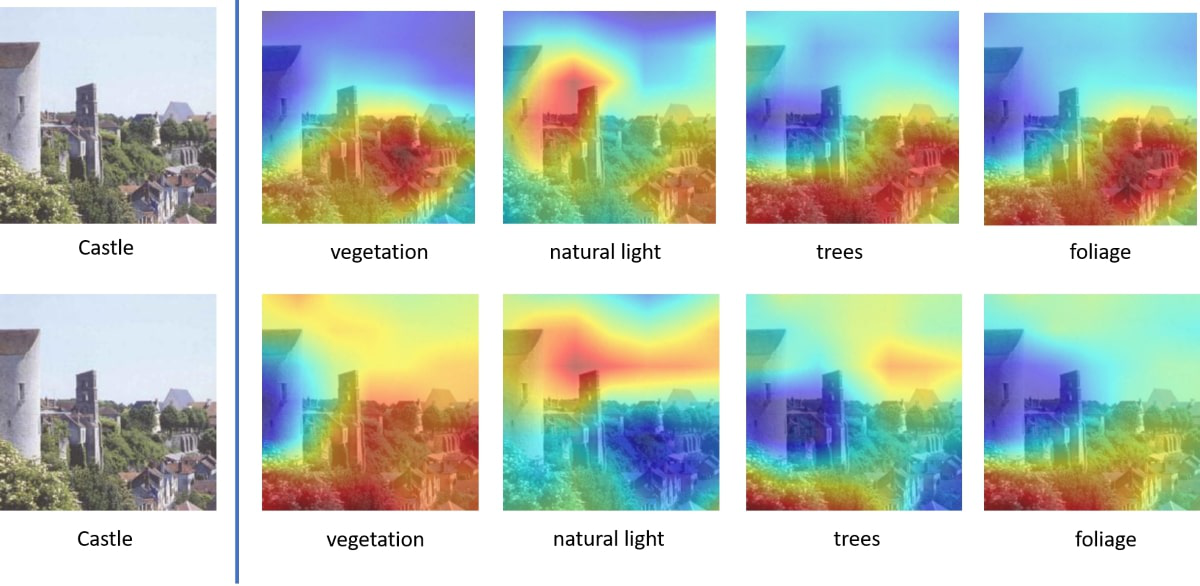}
	\caption{Visualization of attribute maps on SUN from the baseline (top one) and SELAR (bottom one). Below the image are the corresponding attributes.\vspace{-4mm}}
	\label{am_sun}
	\vspace{-2mm}
\end{figure}

\begin{figure}[tb]\centering
	\includegraphics[width=0.50\textwidth,height=0.13\textheight]{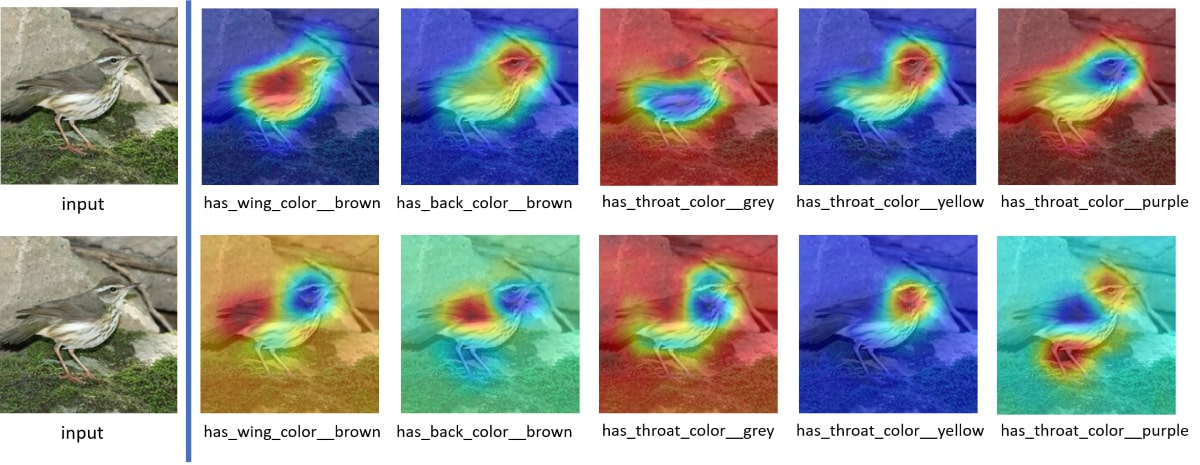}
	\caption{Visualization of attribute maps on CUB from baseline (upper part) and SELAR (lower part). Those attributes have \textbf{lower} value in the attribute vector.\vspace{-2mm}}
	\label{middle_low}
	\vspace{-2mm}
\end{figure}

\minisection{Aggregation method and aggregation space.} We investigate the optimal location to aggregate local features into global ones. The ablation study shown in Table~\ref{table:localization} evaluates GAP and GMP in three different spaces: visual (before the embedding layer), attribute (after embedding layer) and class (after attribute mapping), which also correspond to the order in which features are mapped to the different spaces in our classification pipeline. Note that GAP in either of the three spaces is equivalent due to linearity. Results suggest that the best performance is achieved by GMP in the attribute space.


\minisection{Attribute Activation Maps.} Here we visualize AAMs $\tilde{a}$, for both the baseline and SELAR on SUN in Fig.~\ref{am_sun}. Whereas on the CUB and AWA datasets the attributes are present in clearly localizable small regions, for the SUN dataset this is not the case. This maybe the reason why SELAR and the baseline have similar performance on SUN. While we cannot guarantee that each AAM corresponds to the true attribute, the network learns to correlate the related region with its activation value automatically. We find that AAMs  with high activation values ($>$85) are always highly related to the corresponding attribute. We show additional AAMs with lower attribute values in Fig.~\ref{middle_low}, those attributes have either middle value or do not exist (0). In those cases, the AAM sometimes corresponds to a specific attribute, but sometimes not.

%

\vspace{-2.5mm}
\section{Conclusions}
In this paper, we highlight an overlooked fact in zero-shot learning research, that is, that the feature extractor backbones of common ZSL pipelines can implicitly localize attributes without any explicit region localization module such as attention or part detection. Based on this finding, we propose SELAR, a simple GZSL framework that efficiently extracts highly discriminative representations based on localized attributes. Our method is surprisingly effective, achieving competitive results compared to much more complex state-of-the-art methods. Our findings provide useful insight into how to further design effective and efficient (G)ZSL methods, and advocate SELAR as a strong, yet simple and easy to implement, baseline for future zero-shot learning research.


\bibliographystyle{IEEEtran}
\bibliography{IEEEabrv,egbib}

\end{document}